\documentclass[final]{cvpr}

\usepackage{times}
\usepackage{epsfig}
\usepackage{graphicx}
\usepackage{amsmath}
\usepackage{amssymb}
\usepackage{booktabs}
\usepackage{float}
\usepackage{dblfloatfix}

\usepackage[pagebackref=true,breaklinks=true,colorlinks,bookmarks=false]{hyperref}

\def\cvprPaperID{10} % *** Enter the CVPR Paper ID here
\def\confYear{CVPRW LXCV 2021}

\begin{document}

%%%%%%%%% TITLE
\title{Generalizable Multi-Camera 3D Pedestrian Detection}

\author{João Paulo Lima\textsuperscript{2,1}, Rafael Roberto\textsuperscript{1}, Lucas Figueiredo\textsuperscript{1}, Francisco Simões\textsuperscript{2,1}, Veronica Teichrieb\textsuperscript{1}\\
\textsuperscript{1} Voxar Labs, Centro de Informática, Universidade Federal de Pernambuco\\
\textsuperscript{2} Departamento de Computação, Universidade Federal Rural de Pernambuco\\
{\tt\small \{jpsml, rar3, lsf, fpms, vt\}@cin.ufpe.br, \{joao.mlima,francisco.simoes\}@ufrpe.br}
}

\maketitle

%%%%%%%%% ABSTRACT
\begin{abstract}
   We present a multi-camera 3D pedestrian detection method that does not need to train using data from the target scene.
   We estimate pedestrian location on the ground plane using a novel heuristic based on human body poses and person's bounding boxes from an off-the-shelf monocular detector.
   We then project these locations onto the world ground plane and fuse them with a new formulation of a clique cover problem.
   We also propose an optional step for exploiting pedestrian appearance during fusion by using a domain-generalizable person re-identification model.
   We evaluated the proposed approach on the challenging WILDTRACK dataset.
   It obtained a MODA of 0.569 and an F-score of 0.78, superior to state-of-the-art generalizable detection techniques.
\end{abstract}

%%%%%%%%% BODY TEXT
\section{Introduction}

% why pedestrian detection?
Pedestrian detection is a relevant problem in several contexts, such as smart cities, surveillance, monitoring, autonomous driving, and robotics.
% why 3D?
While several solutions focus only on 2D pedestrian detection~\cite{hasan2020generalizable, liu2019highlevel, liu2020efficient}, estimating the 3D location of pedestrians allows georeferencing them in the environment.
This referencing enables location-based services, spatial visualization, and others~\cite{hackeloeer2013}.
% why multi-camera (challenges: crowded, occlusions)?
Nowadays, it is common that environments have multiple monocular cameras with overlapping fields of view, such as security cameras.
Using such a setup makes 3D pedestrian detection easier since it can exploit multi-view constraints and better handle occlusions.
Nevertheless, multi-camera 3D pedestrian detection in crowded environments is still a challenging task.

% why generalizable?
The methods that currently obtain the best results for detecting pedestrians in 3D using multiple cameras need to perform training using data from the target scene~\cite{hou2020multiview, baque2017deep}.
This implies that they need to retrain when the target scene changes, with different multi-camera configurations and environment conditions.
The training procedure is usually time-demanding and may require laborious annotation of ground-truth data.
Due to this, it is desirable to have a generalizable multi-camera 3D pedestrian detection solution that can be applied out-of-the-box without training with target scene data~\cite{lopez2018semantic, zhu2019multi}.

\begin{figure*}
    \centerline{\includegraphics[width=\textwidth]{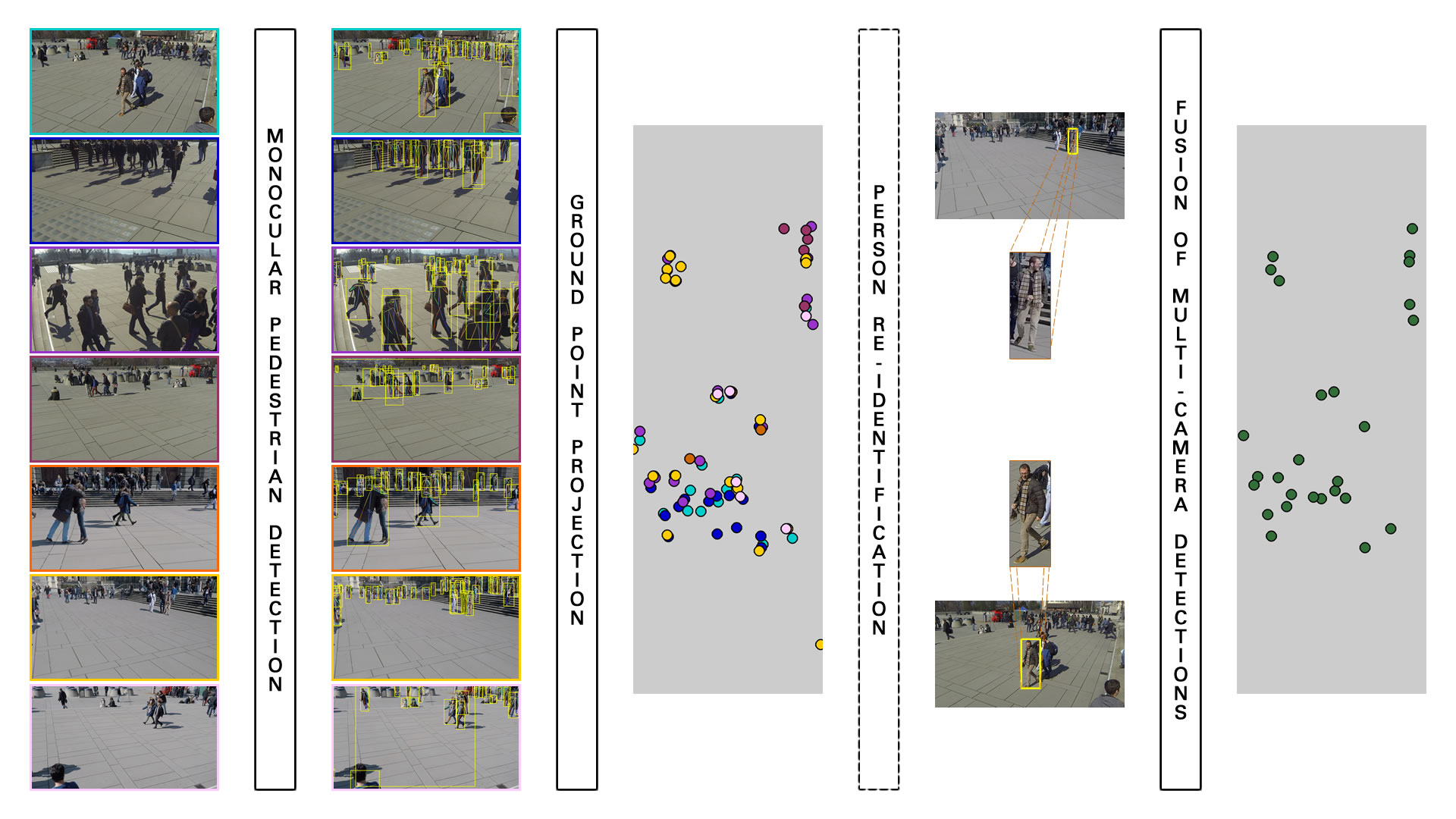}}
    \caption{Summary of our generalizable multi-camera 3D pedestrian detection.
    We detect pedestrians' bounding boxes and body poses in each camera's image (represented by different colors).
    Then we estimate every detection's ground point on every camera and eliminate those out of the area of interest (grey rectangle).
    We can compute a descriptor for each pedestrian bounding box aiming at person re-identification.
    Finally, we solve an instance of the clique cover problem from graph theory to fuse pedestrian detections.}
    \label{fig::summary}
\end{figure*}

% goals
In this context, we present a novel method for multi-camera 3D pedestrian detection classified as generalizable.
The proposed approach encompasses monocular pedestrian detection, estimation of pedestrian location on the ground plane, fusion of multi-camera pedestrian detections, and person re-identification (re-ID).
We summarize our approach in Figure~\ref{fig::summary}.
Since it is based on off-the-shelf monocular person detectors and person re-ID models, it does not require training using target scene data.

% contributions
The contributions of this work are:
\begin{enumerate}
    \item An approach for estimating pedestrian location on the ground plane from off-the-shelf monocular person detectors that provide both human body poses and bounding boxes (Section~\ref{sec::monocular});
    \item A technique for fusing multi-camera pedestrian locations by solving an instance of the clique cover problem from graph theory (Section~\ref{sec::fusion});
    \item An alternative to take pedestrian appearance into account in this fusion step based on domain-generalizable person re-ID models (Section~\ref{sec::reid});
    \item Quantitative and qualitative evaluations regarding the proposed method's detection performance with respect to state-of-the-art generalizable multi-camera 3D detection approaches (Section~\ref{sec::experiments}).
\end{enumerate}

\section{Related Work}

\paragraph{Monocular 3D Pedestrian Detection.} 
3D monocular detectors can retrieve pedestrians' 3D location, requiring a single camera.
MonoPair~\cite{chen2020monopair} uses trained networks to acquire 3D bounding boxes of detected people (among other targets).
It then adds pairwise spatial relationships (based on predicted constraints related to the mid keypoint between targets) to improve the resulting location.
MonoLoco~\cite{bertoni2019monoloco} adopts a monocular approach that infers the depth of each person's detected 2D bounding box based on uncertainty estimation.
It uses 2D estimated poses to model the ambiguity of the 3D location related to intrinsic characteristics such as the different heights of the tracked population.
Hayakawa and Dariush~\cite{hayakawa2020recognition} improve MonoLoco's 3D localization approach by introducing an asymmetric loss function.
It better handles the pixel's related error of estimated joints from distant pedestrians, achieving improved accuracy in these cases.

%retrieve 3D bounding. However, these techniques  The estimation of the 3D location of pedestrians 

%Another approach is to apply 3D multi-view techniques to directly estimate the 3D pedestrian locations \cite{baque2017deep, hou2020multiview, you2020real}.

% falar um pouco (um parágrafo) dos principais resultados dessas técnicas concorrentes, talvez mencionar o que fazemos diferente ou melhoramos sobre elas... \cite{lopez2018semantic, zhu2019multi, ong2020bayesian}

%

% non-generalizable multi-camera 3D pedestrian detection
\paragraph{Non-Generalizable Multi-View 3D Pedestrian Detection.}
In contrast with monocular approaches, estimation using multiple cameras has better results in general because it is prone to deal with complex occlusion problems~\cite{baque2017deep}.
There are multi-view techniques that consider an additional training (supervised or unsupervised) step with the target scene content to better handle contextual information from the application domain.
Given the implicit cost of performing training for each new scene, we can classify them as non-generalizable.
In this sense, Baqué et al.~\cite{baque2017deep} present a combination of convolutional neural networks (CNNs) and conditional random fields (CRFs) to handle the pairwise matching ambiguities between the observed pedestrians.
Alternatively, MVDet~\cite{hou2020multiview} aggregates the multi-view people detection information by applying a feature perspective transform to place all ground heatmaps (and later locations) of pedestrians in the same coordinate space.
Similarly, DMCT~\cite{you2020real} proposes a perspective-aware network, which produces distorted detection blobs (related to the camera's perspective).
This is followed by a fusion procedure for ground-plane occupancy heatmap estimation and the use of a Deep Glimpse Network for person detection.

% generalizable multi-camera 3D pedestrian detection
\paragraph{Generalizable Multi-View 3D Pedestrian Detection.} 
The pipeline to estimate the 3D location of pedestrians in multi-camera scenarios, in a generalizable manner, often employs 2D monocular pedestrian detectors~\cite{hasan2020generalizable, liu2019highlevel, liu2020efficient} and later fuse their results based on multi-view properties~\cite{lopez2018semantic, zhu2019multi, ong2020bayesian}.
In this scenario, one way to make 3D pedestrian detection robust to domain shift, therefore generalizable, is to use monocular person detectors that do not need retraining for a specific target domain~\cite{ren2017faster, he2017mask, li2019crowdpose, hasan2020generalizable}.
Another advantage of monocular detectors re-use is to simplify setup requirements, easing cameras addition/removal/combination~\cite{ong2020bayesian}. 

\section{Monocular Pedestrian Detection and Ground Point Estimation}
\label{sec::monocular}

Our approach relies on an off-the-shelf monocular pedestrian detector that is not retrained for the target domain.
From the monocular detections, we estimate each person's ground point, which is its location on the ground plane.
We can then fuse these ground points for estimating the 3D coordinates of pedestrians in the world ground plane.

Some monocular person detectors can provide both bounding boxes and human body poses as output~\cite{li2019crowdpose}.
The use of full-body poses allows better handling of occlusions, making it possible to verify which parts of the pedestrian's body are visible.
A widely used representation for a body pose is the one employed by the MSCOCO dataset~\cite{lin2014microsoft}, which consists of 17 keypoints that correspond to human body landmarks.
We propose a heuristic for estimating ground points from body poses in the MSCOCO format together with person's bounding boxes.

\begin{figure}[!ht]
    \centerline{\includegraphics[width=31mm]{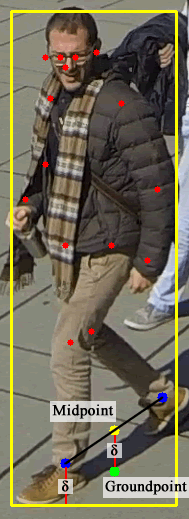}}
    \caption{Midpoint (in yellow) of the two ankles (in blue).
    We find the estimated ground point (in green) by adding the $\delta$ distance to the the midpoint.}
    \label{fig::ground_plane}
\end{figure}

We only use the ankle keypoints, which are the lower ones among all 17 keypoints.
We only keep detections that have scores for ankle keypoints both higher than a threshold~$t_s$.
However, the ankles are not on the ground plane, so we need to apply an offset $\delta$ to the $y$ coordinate of the ankle keypoints to obtain their correspondences on the ground.
This offset is given by

\begin{equation}
    \delta = bb_{y_{max}} - max(la_y, ra_y),
\end{equation}
where $bb_{y_{max}}$ is the $y$ coordinate of the bottom edge of the full-body bounding box, and $la_y$ and $ra_y$ are the $y$ coordinates of left and right ankle keypoints, respectively.
We estimate the ground point as the midpoint of the line segment whose endpoints are the offsetted ankle keypoints.
We illustrate the proposed ground point estimation heuristic in Figure~\ref{fig::ground_plane}.

\section{Fusion of Multi-Camera Detections}
\label{sec::fusion}

Assuming that the cameras are calibrated, the camera frames are undistorted, and that the ground plane corresponds to the $Z=0$ plane in world coordinates, we compute for each camera a homography $\textbf{H}$ that maps the image plane onto the world ground plane.
Considering a camera with intrinsic parameters matrix $\textbf{K}$ and extrinsic parameters matrix $[\textbf{R}|\textbf{t}]$, the projection of the world ground point $\textbf{M}=(X, Y, 0)^T$ onto the image ground point $\textbf{m}=(x, y)^T$ is given by

\begin{equation}
\begin{split}
s
\begin{pmatrix}
x\\ 
y\\ 
1
\end{pmatrix} & = \textbf{K} [\textbf{R}^1 \textbf{R}^2 \textbf{R}^3 \textbf{t}]
\begin{pmatrix}
X\\ 
Y\\ 
0\\
1
\end{pmatrix} \\
& = \textbf{K} [\textbf{R}^1 \textbf{R}^2 \textbf{t}]
\begin{pmatrix}
X\\ 
Y\\
1
\end{pmatrix} \\
& = \textbf{H}^{-1}
\begin{pmatrix}
X\\ 
Y\\
1
\end{pmatrix},
\end{split}
\end{equation}
where $\textbf{R}^i$ is the $i$-th column of \textbf{R}.
Then we use the computed homographies to project all ground points from all cameras to the world ground plane.
If there is a predefined area of interest in the world ground plane, we can use it to discard world ground points outside this area. 

We adopt the same two conditions used by López-Cifuentes et al.~\cite{lopez2018semantic} for fusing world ground points.
All world ground points that correspond to the same pedestrian should have the following properties:

\begin{enumerate}
  \item Come from different cameras since a pedestrian can only appear once per camera frame;
  \item Pairwise Euclidean distances lower than a threshold $t_g$.
\end{enumerate}

As López-Cifuentes et al.~\cite{lopez2018semantic}, we create a graph representing these conditions.
The vertices are the world ground points, and the edges connect vertices associated with world ground points that satisfy both constraints.

We propose to formulate the problem of finding world ground points to be fused as a clique cover problem\footnote{\url{https://en.wikipedia.org/wiki/Clique_cover}}.
A clique is a subset of vertices of a graph such that every two distinct vertices are connected.
Vertices of our graph that belong to a clique represent world ground points that can be fused.
A clique cover is a partition of the vertices of the graph into cliques.
A minimum clique cover is a clique cover that uses as few cliques as possible.
The clique cover problem is the problem of finding a minimum clique cover of a graph.
A clique cover of a graph $G$ may be seen as a graph coloring of the complement graph of $G$.

\begin{figure}[!ht]
    \centerline{\includegraphics[width=\columnwidth]{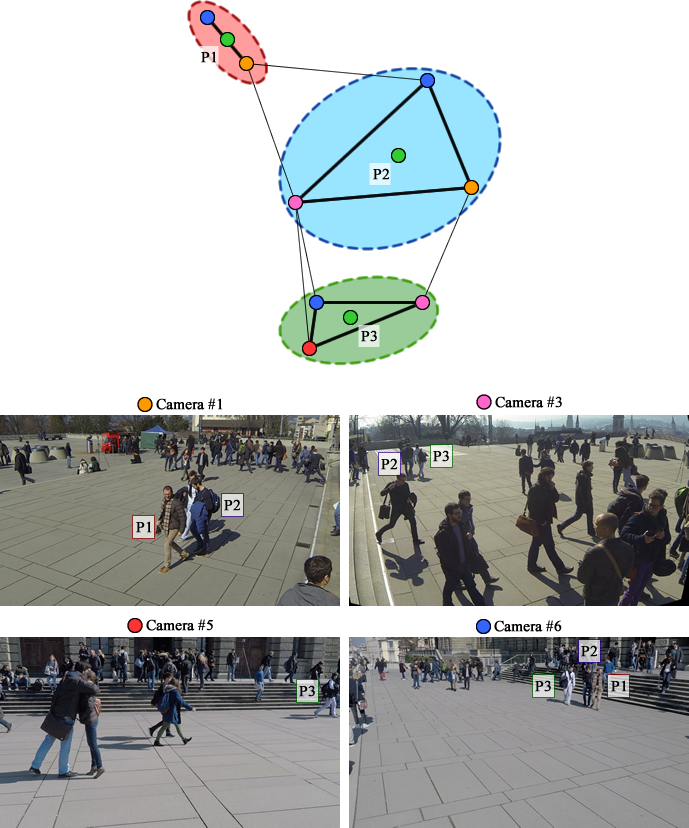}}
    \caption{Graph illustrating the detection of three persons.
    Each node represents the detection by one camera, which can be identified by its color.
    The strong edges in the graph denote different cliques, which we use to compute the person position (green circles over the graph).}
    \label{fig::graph}
\end{figure}

We use the greedy coloring algorithm~\cite{kosowski2004classical} for coloring the complement of the graph created from the world ground points.
We employ the smallest-last vertex ordering strategy with color interchange described by Kosowski and Manuszewski~\cite{kosowski2004classical}.
This prioritizes the fusion of world ground points belonging to pedestrians seen by many cameras at the same time.
Vertices assigned with the same color represent world ground points that meet the fusion criteria.
We assume that the cameras have overlapping fields of view, so we discard cliques with only one vertex.
This helps to decrease the number of false positives since most of the times a pedestrian should appear in more than one camera.
The final 3D coordinate of a detected pedestrian is the arithmetic mean of all world ground points represented by a clique found.
Figure~\ref{fig::graph} illustrates the fusion procedure.

\section{Person Re-Identification}
\label{sec::reid}

So far, we only used the proximity of world ground points for fusing them.
As an optional step, we can also exploit appearance cues for guiding the fusion process.

One way to represent pedestrian appearance is by computing discriminative descriptors invariant to viewing direction and background conditions.
The person re-ID task tackles this problem.
However, person re-ID models often suffer from the domain shift problem, which means that a person re-ID model trained on one source dataset usually presents a degraded performance on an unseen target dataset.
In order to keep multi-camera 3D pedestrian detection generalizable, we need to use a domain-generalizable person re-ID model such as OSNet-IAP~\cite{sovrasov2020building} and OSNet-AIN~\cite{zhou2019learning}.
These models better handle the domain shift issue so that, once trained, they can be deployed without any retraining.

We propose to also use descriptors provided by a domain-generalizable person re-ID model to help the fusion of world ground points.
The person re-ID model takes as input the pedestrian bounding box cropped out of the respective camera frame and outputs a high-dimensional vector as a descriptor.
We can then have an additional condition that all world ground points belonging to the same pedestrian should satisfy:

\begin{enumerate}
  \setcounter{enumi}{2}
  \item Pairwise descriptor distances lower than a threshold $t_d$.
\end{enumerate}

When using person re-ID, we should take into account all the three criteria presented for creating the fusion graph (following the approach explained in Section~\ref{sec::fusion}).

\section{Experiments}
\label{sec::experiments}

We evaluated the proposed method under a multi-camera 3D pedestrian detection scenario with a crowded setup.
We present in the following subsections the details of the experiments carried out and the results obtained.

\subsection{Dataset and Metrics}

We used the publicly available and challenging WILDTRACK dataset\footnote{\url{https://www.epfl.ch/labs/cvlab/data/data-wildtrack/}}~\cite{chavdarova2018wildtrack}, which was acquired using seven static cameras with overlapping fields of view in a crowded public open area.
It provides both intrinsic and extrinsic calibration for each camera and synchronized frames with a $1920\times1080$ resolution.
Ground-truth 3D locations of pedestrians are available for 400 frames at 2 fps, covering an area of interest of $12\times36$ m, with an average of 23.8 people per frame and a total of 9518 annotations.

We followed the evaluation protocol proposed by Chavdarova et al.~\cite{chavdarova2018wildtrack}, which uses the subsequent metrics: Multiple Object Detection Accuracy (MODA), Multiple Object Detection Precision (MODP), precision (Prcn), and recall (Rcll). The 3D detections are assigned to ground truth using Hungarian matching and only if they are closer than $0.5 m$.
We consider MODA as the primary performance indicator since it takes into account both false negatives and false positives.

\subsection{Environment Setup}

The hardware used in the evaluations was a laptop with an Intel Core i7-7700HQ @2.80 GHz processor, 32 GB RAM, and an NVIDIA GeForce GTX 1060 graphics card.

We used AlphaPose\footnote{\url{https://github.com/MVIG-SJTU/AlphaPose}}~\cite{li2019crowdpose} for human body pose estimation, which employs YOLOv3\footnote{\url{https://pjreddie.com/darknet/yolo/}}~\cite{redmon2018yolov3} for person bounding box detection.
We performed greedy graph coloring with the algorithm implementation available in the NetworkX\footnote{\url{https://networkx.org/}} package for Python.
For person re-ID, we used an OpenVINO\footnote{\url{https://docs.openvinotoolkit.org/}} model based on the omni-scale network (OSNet) backbone with Linear Context Transform (LCT) blocks~\cite{sovrasov2020building}.
It outputs 256-dimensional descriptors that we compare using the cosine distance.

In all experiments we used empirically obtained threshold values for the proposed method as following: keypoint score threshold $t_s=0.4$, ground point distance threshold $t_g=0.7 m$ and descriptor distance threshold $t_d=1.0$.
We report the results of competing approaches using the optimal parameter values found.

\subsection{Detection Performance Evaluation}

First, we evaluated different approaches for monocular pedestrian detection and ground point estimation:

\begin{itemize}
    \item \textbf{BB only}, which uses only the bounding boxes provided by the YOLOv3 detector and estimates ground points as Zhu~\cite{zhu2019multi} and López-Cifuentes et al.~\cite{lopez2018semantic} by taking the center of the bottom edge of the bounding boxes;
    \item \textbf{BP \& BB}, which is the proposed approach described in Section~\ref{sec::monocular} based on both body poses and bounding boxes.
\end{itemize}

\begin{figure}[h]
    \centerline{\includegraphics[width=\columnwidth]{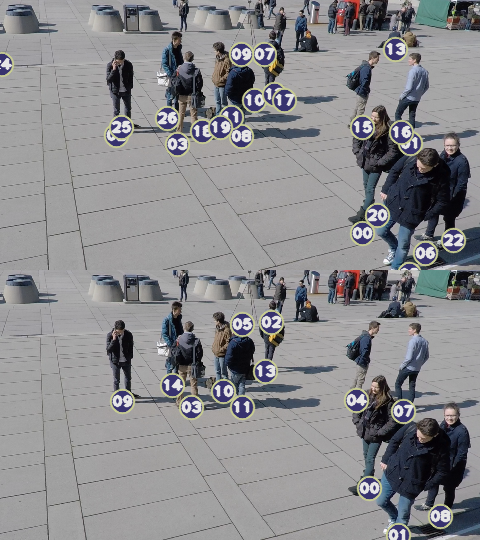}}
    \caption{3D detections projected onto frame \#1650 from camera~7 of the WILDTRACK dataset.
    Top: BB only + CC.
    Bottom: BP \& BB + CC.}
    \label{fig::gpe_qual}
\end{figure}

We tested both strategies together with the proposed approach detailed in Section~\ref{sec::fusion} for the fusion of multi-camera detections based on the clique cover problem (CC).
As can be seen in Table~\ref{tab::monocular}, the proposed BP \& BB method presented significantly better results than the BB only approach.
Figure~\ref{fig::gpe_qual} shows examples of results obtained using BB only + CC and BP \& BB + CC.
The proposed BP \& BB approach presented far fewer false positives than the BB only method for this given frame.

% comparison of monocular pedestrian detectors and ground points estimators (AlphaPose vs YOLO)
\begin{table}[h]
\centering
\begin{tabular}{@{}lrrrr@{}}
\toprule
Method         & \multicolumn{1}{c}{MODA} & \multicolumn{1}{c}{MODP} & \multicolumn{1}{c}{Prcn} & \multicolumn{1}{c}{Rcll} \\ \midrule
BB only + CC  & 0.147                      & 0.587                     & 0.560                          & 0.689                       \\
BP \& BB + CC & \textbf{0.569}             & \textbf{0.673}            & \textbf{0.808}                 & \textbf{0.746}              \\ \bottomrule
\end{tabular}
\caption{\label{tab::monocular}Performance evaluation of different strategies for monocular pedestrian detection and ground point estimation: using only bounding boxes (BB only) and the proposed approach using both body poses and bounding boxes (BP \& BB).
We employ the proposed clique cover method (CC) for the fusion of multi-camera detections in both strategies.}
\end{table}

In the next experiment, we compared using the BP \& BB method together with different strategies for the fusion of multi-camera detections:

\begin{itemize}
    \item \textbf{AH}, which computes an average heatmap as described by You and Jiang~\cite{you2020real}.
    It obtains each camera's heatmaps by considering a non-normalized Gaussian kernel in world ground plane coordinates centered at each ground point with a radius of $0.8 m$ and $\sigma=10.1$.
    Then it retrieves the detections as local maxima on the average heatmap with a minimum allowed distance of $0.5 m$ and minimum value of $0.3$;
    \item \textbf{GH}, which is the greedy heuristic presented by Zhu~\cite{zhu2019multi} using a distance of $0.8 m$ for combining detections;
    \item \textbf{CC}, which is the proposed approach detailed in Section~\ref{sec::fusion} based on the clique cover problem.
\end{itemize}

\begin{figure}[h]
    \centerline{\includegraphics[width=\columnwidth]{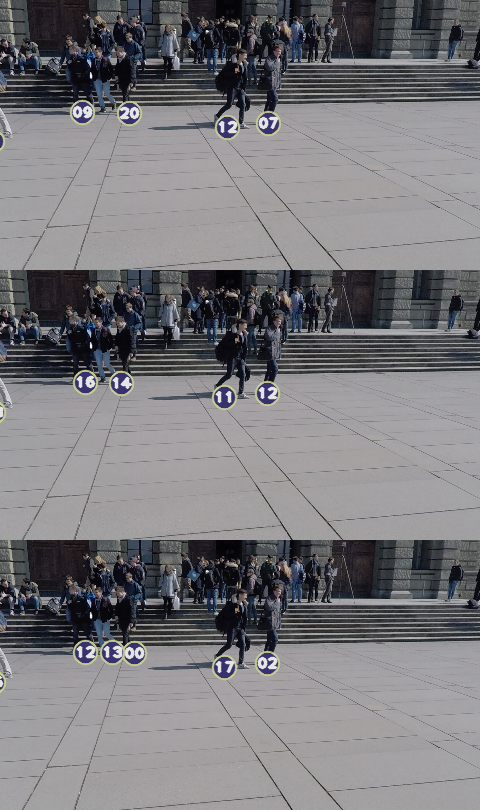}}
    \caption{3D detections projected onto frame \#340 from camera~5 of the WILDTRACK dataset.
    Top: BP \& BB + AH.
    Middle: BP \& BB + GH.
    Bottom: BP \& BB + CC.}
    \label{fig::fusion_qual}
\end{figure}

Table~\ref{tab::fusion} shows that the proposed CC method obtained the best results with respect to MODA, MODP, and precision.
The recall obtained by CC was higher than by AH and slightly lower than by GH.
Figure~\ref{fig::fusion_qual} depicts examples of results obtained using BP \& BB + AH, BP \& BB + GH and BP \& BB + CC.
While one false negative appears in the presented view for both AH and GH, the proposed CC approach correctly detects all persons visible in this view that are inside the area of interest.

% comparison of ground point fusion methods (clique cover vs greedy method from Zhu 2019 vs average occupancy map from You & Jiang 2020)
\begin{table}[h]
\centering
\begin{tabular}{@{}lrrrr@{}}
\toprule
Method          & \multicolumn{1}{c}{MODA} & \multicolumn{1}{c}{MODP} & \multicolumn{1}{c}{Prcn} & \multicolumn{1}{c}{Rcll} \\ \midrule
BP \& BB + AH  & 0.262                     & 0.670            & 0.608                          & 0.736                       \\
BP \& BB + GH  & 0.564                     & 0.670            & 0.802                          & \textbf{0.749}              \\
BP \& BB + CC  & \textbf{0.569}            & \textbf{0.673}                     & \textbf{0.808}                 & 0.746                       \\ \bottomrule
\end{tabular}
\caption{\label{tab::fusion}Performance evaluation of different strategies for fusion of multi-camera detections: using an average heatmap (AH), a greedy heuristic (GH) and the proposed approach based on the clique cover problem (CC).
In all strategies, we employ the proposed method for monocular pedestrian detection and ground point estimation using both body poses and bounding boxes (BP \& BB).}
\end{table}

\begin{figure}[h]
    \centerline{\includegraphics[width=\columnwidth]{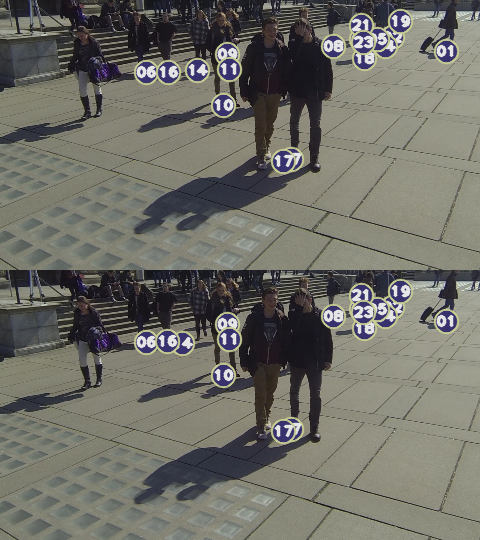}}
    \caption{3D detections projected onto frame \#800 from camera~2 of the WILDTRACK dataset.
    Top: BP \& BB + CC.
    Bottom: BP \& BB + CC + Re-ID.}
    \label{fig::reid_qual}
\end{figure}

We also evaluated the effect of using person re-ID as proposed in Section~\ref{sec::reid}.
The addition of person re-ID brought almost no changes to MODA, MODP, precision, and recall.
Due to this, we show in Table~\ref{tab::reid} the number of true positives, false positives, and false negatives.
It is worth noting that using person re-ID caused a slight increase of false positives.
Figure~\ref{fig::reid_qual} depicts examples of results obtained using BP \& BB + CC and BP \& BB + CC + Re-ID.
The approach that employs re-ID could not correctly connect all graph nodes belonging to the same person.
This caused a noticeable shift in the location of detection \#14, resulting in one false negative and one false positive.

% evaluation of the use of person re-ID (with vs without)
\begin{table}[h]
\centering
\begin{tabular}{@{}lrrr@{}}
\toprule
Method                & \multicolumn{1}{c}{TP} & \multicolumn{1}{c}{FP} & \multicolumn{1}{c}{FN} \\ \midrule
BP \& BB + CC         & 7096                   & \textbf{1683}          & 2422                   \\
BP \& BB + CC + Re-ID & 7096                   & 1685                   & 2422                   \\ \bottomrule
\end{tabular}
\caption{\label{tab::reid}Performance evaluation of the proposed multi-camera 3D pedestrian detection method without (BP \& BB + CC) and with (BP \& BB + CC + Re-ID) the proposed person re-ID approach with respect to number of true positives (TP), false positives (FP) and false negatives (FN).}
\end{table}

Table~\ref{tab::sota} compares the results obtained with the best configuration of the proposed multi-camera 3D pedestrian detection method to state-of-the-art approaches that can be classified as generalizable.
The results of RCNN-projected, POM-CNN and Pre-DeepMCD are the ones reported by Chavdarova et al.~\cite{chavdarova2018wildtrack}, and the results of López-Cifuentes et al. 2018~\cite{lopez2018semantic} and Zhu 2019~\cite{zhu2019multi} are the ones reported in their respective works.
Since some methods only reported F-score instead of precision and recall, we also added this metric to the evaluation. 
Our technique outperformed all other approaches regarding MODA and F-score.

% comparison with state-of-the-art non-trained approaches
\begin{table*}[!b]
\centering
\begin{tabular}{@{}lccccc@{}}
\toprule
Method						& MODA				& MODP				& Precision	& Recall	& F-Score		\\ \midrule
RCNN-projected				& 0.113				& 0.184				& 0.680		& 0.430		& 0.53			\\
POM-CNN						& 0.232				& 0.305				& 0.750		& 0.550		& 0.63			\\
Pre-DeepMCD					& 0.334				& 0.528				& 0.930		& 0.360		& 0.52			\\
López-Cifuentes et al. 2018	& 0.390				& 0.550				& -			& -			& 0.69			\\
Zhu 2019					& 0.540				& \textbf{0.820}	& -			& -			& 0.77			\\
BP \& BB + CC (ours)		& \textbf{0.569}	& 0.673				& 0.808		& 0.746		& \textbf{0.78}	\\ \bottomrule
\end{tabular}
\caption{\label{tab::sota}Performance comparison of the proposed multi-camera 3D pedestrian detection approach (BP \& BB + CC) with state-of-the-art methods not trained on the target dataset.}
\end{table*}

\begin{figure*}[!ht]
    \centerline{\includegraphics[width=\textwidth]{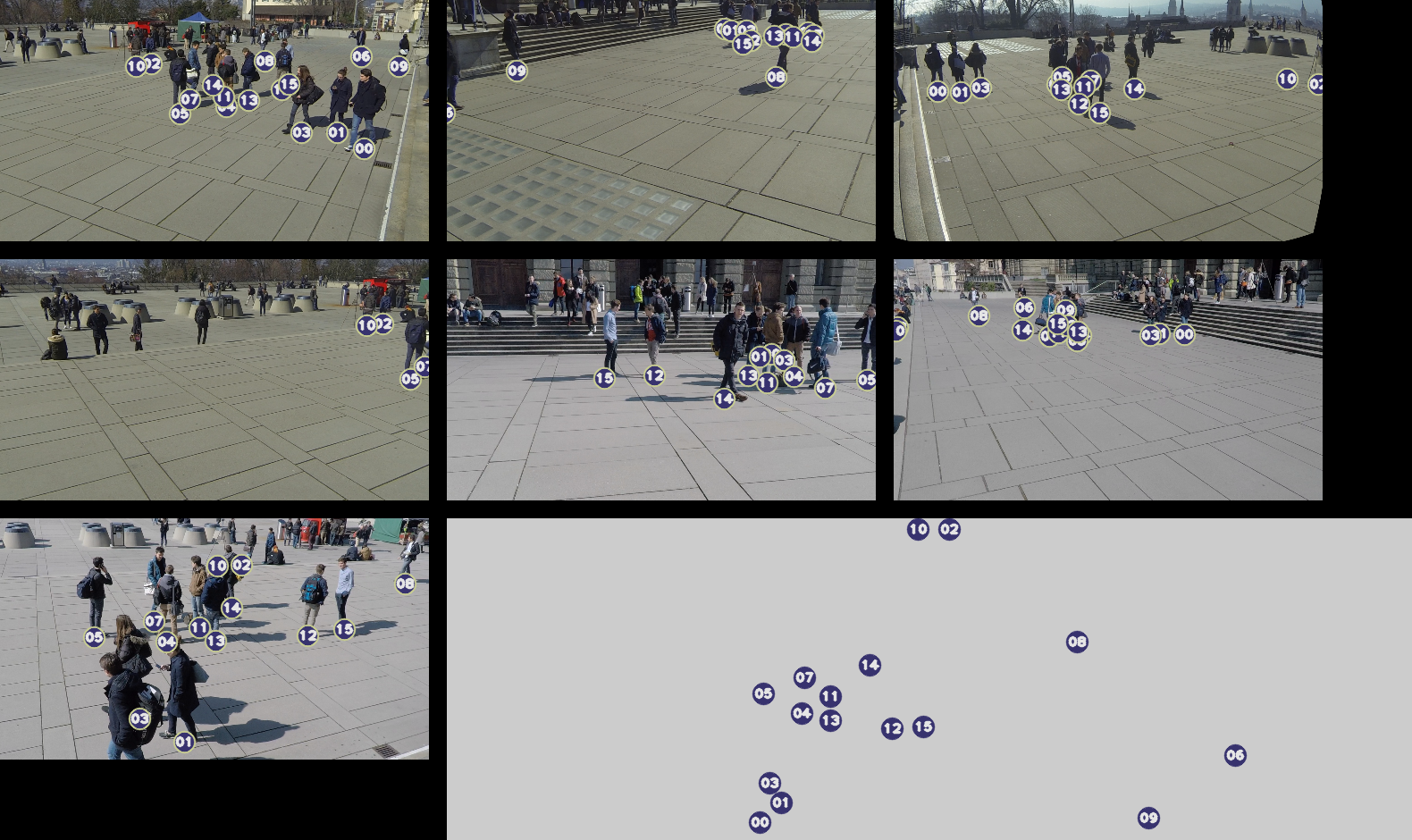}}
    \caption{3D pedestrian detection results obtained with the proposed BP \& BB + CC method for frame \#1685 of the WILDTRACK dataset.
    Blue circles represent detected pedestrians.
    At the bottom right, we show their locations on the world ground plane.
    From left to right, top to bottom, we see the frames from cameras 1 to 7 with the projection of all 3D detections.
    Detections with equal number labels refer to the same pedestrian.}
    \label{fig::qualitative}
\end{figure*}

% qualitative evaluation
Figure~\ref{fig::qualitative} and supplementary material depict a visualization from all views and from the world ground plane of a result obtained using BP \& BB + CC.
We can note that the proposed method could correctly detect the 3D locations of all pedestrians inside the area of interest.

\subsection{Execution Time Analysis}

Table~\ref{tab::time} presents a time performance analysis of a non-optimized version of the proposed approach.
Fusion of multi-camera detections is executed only once per frame, while the other procedures are executed once per camera for each frame.
The bottleneck is the monocular pedestrian detection and ground point estimation step.

\subsection{Limitations}

% restricted to ground plane
Similar to many other existing multi-camera 3D pedestrian detection methods in the literature~\cite{hou2020multiview, you2020real, zhu2019multi, lopez2018semantic, baque2017deep}, our proposed approach is restricted to the ground plane.
Therefore, it cannot correctly estimate the 3D location of people who are not standing on the ground (e.g., jumping).
This may limit its application to domains such as sports analytics.

\begin{figure}[!h]
    \centerline{\includegraphics[width=\columnwidth]{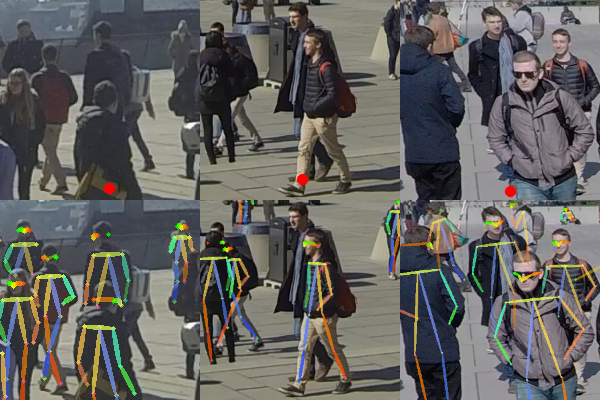}}
    \caption{Failure case of the proposed method due to severe occlusions.
    Top row: patches from views of WILDTRACK frame \#1990 where non-detected pedestrian \#1120 appears, with red circles representing his projected ground truth location.
    Bottom row: body pose estimation results for each respective image.}
    \label{fig::occlusion}
\end{figure}

% severe occlusions
Our method also fails when the pedestrian suffers severe occlusions that prevent ankle keypoints from being reliably detected.
Figure~\ref{fig::occlusion} illustrates such a situation, where AlphaPose was not able to estimate the skeleton of a pedestrian in one of the views and missed the ankle joints in the other two views where he appears.
Since we have no ground point for this pedestrian, we end up with a false negative.

\begin{table*}[h]
\centering
\begin{tabular}{ll}
\hline
Procedure                                                               & \multicolumn{1}{c}{Time (ms)} \\ \hline
Monocular pedestrian detection and ground point estimation (per camera) & \multicolumn{1}{r}{$829.1\pm158.3$}          \\
Ground point projection and area of interest filtering (per camera)     & \multicolumn{1}{r}{$0.1\pm0.3$}          \\
Person re-ID descriptors computation (per camera)                                               & \multicolumn{1}{r}{$186.9\pm67.6$}          \\
Fusion of multi-camera detections (per frame)                           & \multicolumn{1}{r}{$224.0\pm71.3$}          \\ \hline
\end{tabular}
\caption{\label{tab::time}Mean and standard deviation of time spent by each procedure of the proposed method for each frame.}
\end{table*}

\section{Conclusion}

We presented a new approach for multi-camera 3D pedestrian detection that is generalizable, not requiring scene-dependent training.
The proposed method for ground point estimation based on human body poses and bounding boxes proved superior to the commonly used midpoint of the bounding box base.
The novel technique for the fusion of ground points as a clique cover problem obtained better results than existing techniques in the literature.
The suggested use of a domain-generalizable person re-ID model for giving additional cues to ground point fusion did not bring any improvements.
Our approach outperformed state-of-the-art generalizable detection methods.

As future work, we plan to investigate the use of a multi-person 3D pose estimation method~\cite{he2020multi} as an auxiliary for multi-camera 3D pedestrian detection.
We believe this may help to cope with limitations such as restriction to the ground plane and sensitivity to severe occlusions.

\section*{Acknowledgements}
The authors would like to thank Conselho Nacional de Desenvolvimento Cient\'{i}fico e Tecnol\'{o}gico (CNPq) (process 425401/2018-9) for partially funding this research.

{\small
\bibliographystyle{ieee_fullname}
\bibliography{egbib}

\begin{thebibliography}{10}\itemsep=-1pt

\bibitem{baque2017deep}
P. {Baqué}, F. {Fleuret}, and P. {Fua}.
\newblock Deep occlusion reasoning for multi-camera multi-target detection.
\newblock In {\em 2017 IEEE International Conference on Computer Vision
  (ICCV)}, pages 271--279, 2017.

\bibitem{bertoni2019monoloco}
L. {Bertoni}, S. {Kreiss}, and A. {Alahi}.
\newblock Monoloco: Monocular 3d pedestrian localization and uncertainty
  estimation.
\newblock In {\em 2019 IEEE/CVF International Conference on Computer Vision
  (ICCV)}, pages 6860--6870, 2019.

\bibitem{chavdarova2018wildtrack}
T. {Chavdarova}, P. {Baqué}, S. {Bouquet}, A. {Maksai}, C. {Jose}, T.
  {Bagautdinov}, L. {Lettry}, P. {Fua}, L. {Van Gool}, and F. {Fleuret}.
\newblock Wildtrack: A multi-camera hd dataset for dense unscripted pedestrian
  detection.
\newblock In {\em 2018 IEEE/CVF Conference on Computer Vision and Pattern
  Recognition}, pages 5030--5039, 2018.

\bibitem{he2020multi}
He Chen, Pengfei Guo, Pengfei Li, Gim~Hee Lee, and Gregory Chirikjian.
\newblock Multi-person 3d pose estimation in crowded scenes based on multi-view
  geometry.
\newblock In Andrea Vedaldi, Horst Bischof, Thomas Brox, and Jan-Michael Frahm,
  editors, {\em Computer Vision -- ECCV 2020}, pages 541--557, Cham, 2020.
  Springer International Publishing.

\bibitem{chen2020monopair}
Y. {Chen}, L. {Tai}, K. {Sun}, and M. {Li}.
\newblock Monopair: Monocular 3d object detection using pairwise spatial
  relationships.
\newblock In {\em 2020 IEEE/CVF Conference on Computer Vision and Pattern
  Recognition (CVPR)}, pages 12090--12099, 2020.

\bibitem{hackeloeer2013}
Andreas Hackeloeer, Klaas Klasing, Jukka~M. Krisp, and Liqiu Meng.
\newblock Georeferencing: a review of methods and applications.
\newblock {\em Annals of GIS}, 20(1):61--69, 2014.

\bibitem{hasan2020generalizable}
Irtiza Hasan, Shengcai Liao, Jinpeng Li, Saad~Ullah Akram, and Ling Shao.
\newblock Generalizable pedestrian detection: The elephant in the room.
\newblock {\em arXiv preprint arXiv:2003.08799}, 2020.

\bibitem{hayakawa2020recognition}
J. {Hayakawa} and B. {Dariush}.
\newblock Recognition and 3d localization of pedestrian actions from monocular
  video.
\newblock In {\em 2020 IEEE 23rd International Conference on Intelligent
  Transportation Systems (ITSC)}, pages 1--7, 2020.

\bibitem{he2017mask}
K. {He}, G. {Gkioxari}, P. {Dollár}, and R. {Girshick}.
\newblock Mask r-cnn.
\newblock In {\em 2017 IEEE International Conference on Computer Vision
  (ICCV)}, pages 2980--2988, 2017.

\bibitem{hou2020multiview}
Yunzhong Hou, Liang Zheng, and Stephen Gould.
\newblock Multiview detection with feature perspective transformation.
\newblock In Andrea Vedaldi, Horst Bischof, Thomas Brox, and Jan-Michael Frahm,
  editors, {\em Computer Vision -- ECCV 2020}, pages 1--18, Cham, 2020.
  Springer International Publishing.

\bibitem{kosowski2004classical}
Adrian Kosowski and Krzysztof Manuszewski.
\newblock Classical coloring of graphs.
\newblock {\em Contemporary Mathematics}, 352:1--20, 2004.

\bibitem{li2019crowdpose}
J. {Li}, C. {Wang}, H. {Zhu}, Y. {Mao}, H. {Fang}, and C. {Lu}.
\newblock Crowdpose: Efficient crowded scenes pose estimation and a new
  benchmark.
\newblock In {\em 2019 IEEE/CVF Conference on Computer Vision and Pattern
  Recognition (CVPR)}, pages 10855--10864, 2019.

\bibitem{lin2014microsoft}
Tsung-Yi Lin, Michael Maire, Serge Belongie, James Hays, Pietro Perona, Deva
  Ramanan, Piotr Doll{\'a}r, and C.~Lawrence Zitnick.
\newblock Microsoft coco: Common objects in context.
\newblock In David Fleet, Tomas Pajdla, Bernt Schiele, and Tinne Tuytelaars,
  editors, {\em Computer Vision -- ECCV 2014}, pages 740--755, Cham, 2014.
  Springer International Publishing.

\bibitem{liu2020efficient}
W. {Liu}, S. {Liao}, and W. {Hu}.
\newblock Efficient single-stage pedestrian detector by asymptotic localization
  fitting and multi-scale context encoding.
\newblock {\em IEEE Transactions on Image Processing}, 29:1413--1425, 2020.

\bibitem{liu2019highlevel}
W. {Liu}, S. {Liao}, W. {Ren}, W. {Hu}, and Y. {Yu}.
\newblock High-level semantic feature detection: A new perspective for
  pedestrian detection.
\newblock In {\em 2019 IEEE/CVF Conference on Computer Vision and Pattern
  Recognition (CVPR)}, pages 5182--5191, 2019.

\bibitem{lopez2018semantic}
Alejandro L{\'o}pez-Cifuentes, Marcos Escudero-Vi{\~n}olo, Jes{\'u}s
  Besc{\'o}s, and Pablo Carballeira.
\newblock Semantic driven multi-camera pedestrian detection.
\newblock {\em arXiv preprint arXiv:1812.10779}, 2018.

\bibitem{ong2020bayesian}
J. {Ong}, B.~T. {Vo}, B.~N. {Vo}, D.~Y. {Kim}, and S. {Nordholm}.
\newblock A bayesian filter for multi-view 3d multi-object tracking with
  occlusion handling.
\newblock {\em IEEE Transactions on Pattern Analysis and Machine Intelligence},
  pages 1--1, 2020.

\bibitem{redmon2018yolov3}
Joseph Redmon and Ali Farhadi.
\newblock Yolov3: An incremental improvement.
\newblock {\em arXiv preprint arXiv:1804.02767}, 2018.

\bibitem{ren2017faster}
S. {Ren}, K. {He}, R. {Girshick}, and J. {Sun}.
\newblock Faster r-cnn: Towards real-time object detection with region proposal
  networks.
\newblock {\em IEEE Transactions on Pattern Analysis and Machine Intelligence},
  39(6):1137--1149, 2017.

\bibitem{sovrasov2020building}
Vladislav Sovrasov and Dmitry Sidnev.
\newblock Building computationally efficient and well-generalizing person
  re-identification models with metric learning.
\newblock {\em arXiv preprint arXiv:2003.07618}, 2020.

\bibitem{you2020real}
Quanzeng You and Hao Jiang.
\newblock Real-time 3d deep multi-camera tracking.
\newblock {\em arXiv preprint arXiv:2003.11753}, 2020.

\bibitem{zhou2019learning}
Kaiyang Zhou, Yongxin Yang, Andrea Cavallaro, and Tao Xiang.
\newblock Learning generalisable omni-scale representations for person
  re-identification.
\newblock {\em arXiv preprint arXiv:1910.06827}, 2019.

\bibitem{zhu2019multi}
Chuting Zhu.
\newblock Multi-camera people detection and tracking, 2019.

\end{thebibliography}
}

\end{document}